\documentclass[10pt,journal]{IEEEtran}
\usepackage{cite}
\ifCLASSINFOpdf
  \usepackage[pdftex]{graphicx}
\else
  \usepackage[dvips]{graphicx}
\fi
\usepackage{amsmath}
\usepackage{amsthm}
\usepackage{algorithmic}
\usepackage{array}
\usepackage{tikz}
\usepackage{multirow}
\usetikzlibrary{decorations.pathreplacing}
\ifCLASSOPTIONcompsoc
  \usepackage[caption=false,font=normalsize,labelfont=sf,textfont=sf]{subfig}
\else
  \usepackage[caption=false,font=footnotesize]{subfig}
\fi
\usepackage{fixltx2e}

\ifCLASSOPTIONcaptionsoff
  \usepackage[nomarkers]{endfloat}
 \let\MYoriglatexcaption\caption
 \renewcommand{\caption}[2][\relax]{\MYoriglatexcaption[#2]{#2}}
\fi
\usepackage{url}
\usepackage{amssymb}
\usepackage{bbm}
\usepackage{cleveref}
\usepackage{fancyhdr}

\newtheorem*{definition}{Definition}

\fancypagestyle{copyright}{
\fancyhf{}

\fancyfoot[L]{\footnotesize \vspace{-1.5\baselineskip} \copyright 2021 IEEE. Personal use of this material is permitted.  Permission from IEEE must be obtained for all other uses, in any current or future media, including reprinting/republishing this material for advertising or promotional purposes, creating new collective works, for resale or redistribution to servers or lists, or reuse of any copyrighted component of this work in other works.}
}
\begin{document}
\title{Metaparametric Neural Networks for Survival Analysis}

\author{Fabio Luis~de Mello,
        J Mark~Wilkinson,
	and~Visakan~Kadirkamanathan
\thanks{F.L. de Mello was funded by a National Joint Registry
studentship.}
\thanks{F.L. de Mello, J.M. Wilkinson and V. Kadirkamanathan are with the University of Sheffield, S10 2TN Sheffield, UK (e-mail: fldemello1@sheffield.ac.uk; j.m.wilkinson@sheffield.ac.uk; visakan@sheffield.ac.uk)}}

\fancyfoot[L]{\copyright 2021 IEEE.  Personal use of this material is permitted.  Permission from IEEE must be obtained for all other uses, in any current or future media, including reprinting/republishing this material for advertising or promotional purposes, creating new collective works, for resale or redistribution to servers or lists, or reuse of any copyrighted component of this work in other works.}

\maketitle

\thispagestyle{copyright}

\begin{abstract}
Survival analysis is a critical tool for the modelling of time-to-event data, such as life expectancy after a cancer diagnosis or optimal maintenance scheduling for complex machinery. However, current neural network models provide an imperfect solution for survival analysis as they either restrict the shape of the target probability distribution or restrict the estimation to pre-determined times. As a consequence, current survival neural networks lack the ability to estimate a generic function without prior knowledge of its structure. In this article, we present the metaparametric neural network framework that encompasses existing survival analysis methods and enables their extension to solve the aforementioned issues. This framework allows survival neural networks to satisfy the same independence of generic function estimation from the underlying data structure that characterizes their regression and classification counterparts. Further, we demonstrate the application of the metaparametric framework using both simulated and large real-world datasets and show that it outperforms the current state-of-the-art methods in (i) capturing nonlinearities, and (ii) identifying temporal patterns, leading to more accurate overall estimations whilst placing no restrictions on the underlying function structure.
\end{abstract}


\IEEEpeerreviewmaketitle

\section{Introduction}
\IEEEPARstart{S}{urvival} analysis models estimate how the probability of one or more events evolve with time, depending upon a given set of input attributes. Survival models find wide application across society. For example, in healthcare they are used to estimate the influence of risk factors upon disease
\cite{bib:prospective2002age, bib:clift2020living};
the effectiveness of vaccines \cite{bib:vasileiou2021effectiveness}
and medications \cite{bib:pitt1999effect};
and their associated risks
\cite{bib:haaberg2013risk, bib:aram2018estimating}.
In economics, applications include the modelling of unemployment duration \cite{bib:kiefer1988economic} and the detection of financial misconduct in stock markets \cite{bib:karpoff2010short}.
In industry, applications include the estimation of remaining useful lifetime for machinery \cite{bib:sikorska2011prognostic}, and in business to inform our understanding of the timeframes over which new technologies are adopted \cite{bib:liu2013time}.

A key challenge in survival analysis that distinguishes it from regression or
classification problems is the requirement to estimate the probability of an
event over time in the presence of censored data.
Within a given timeframe, the event of interest may
not occur and is said to be censored.
The task of the model is to use all the available data to estimate the event
probability at any given time, as a function of the input variables.
The incorporation of censored data in the model is critical to avoid estimation
bias, and adds to the complexity of survival analysis.
Traditional solutions to survival estimation include semi-parametric or
parametric models that rely on assumptions about the structure of the survival
probability distribution \cite{bib:cox1972regression, bib:royston2002flexible}.
	
Given the broad range of practical applications, there is substantial interest
in applying machine learning techniques, and neural networks in particular, to
solve survival analysis problems.
This endeavour is inspired by previous success in regression and classification
tasks, in which almost any scenario can be modelled generically without prior
knowledge of the underlying functional relationships.
In these tasks, the network output can be achieved with the appropriate choice
of activation function, such as linear activation in regression
\cite{bib:tang2015extreme}, sigmoid activation in Boolean classification
\cite{bib:bianchini2014complexity} or softmax activation in multinomial
classification \cite{bib:zhao2019object}.
Other tools may also be applied within the generalized framework to improve the
extraction of hidden features from the input data, such as convolutional neural
networks in the case of images \cite{bib:zhao2019object} and long short-term
memory for time series analysis \cite{bib:greff2016lstm}.
However, this neural network functional representation cannot be used directly
to represent time and input-dependent probability distribution functions.
Similarly, although current survival analysis methodologies allow feature extraction from the inputs, the target distribution is not generically
parameterized as a function of those features.

Currently, one of two frameworks are typically used for subject-specific
survival analysis: the proportional hazards model \cite{bib:cox1972regression}
or the accelerated failure time (AFT) model \cite{bib:cox1972regression,
	bib:miller1976least,bib:buckley1979linear,bib:peng2008survival}.
The proportional hazards model is built upon the assumption that the
instantaneous probability of an event, i.e. the hazard function, has a baseline
time structure that is similar for all subjects.
This structure can be amplified or attenuated by a factor that depends on the
covariates, but is time-constant.
If one of the input covariates depends on time or produces an effect on the
hazard function that is time-dependent, this can be taken into account using
basis functions \cite{bib:moreau1985global, bib:gray1992flexible,
bib:hastie1993varying,bib:sauerbrei2006multivariable}. In some problems more
than one type of event may occur.
These are termed competing risks scenarios.
Competing risks extensions of the model have been proposed by
\cite{bib:prentice1978analysis,bib:larson1985mixture,bib:fine1999proportional}.
The AFT model is also based on the assumption of a baseline time structure that
is common to all subjects.
Here, instead of amplifying or attenuating the risk of an event, the covariates
accelerate or decelerate the failure process.
A competing risks extension of the AFT model has also been proposed
\cite{bib:choi2018semiparametric}.
Early AFT models required the baseline hazard function to be constrained to a
specific family of functions, making it less generic than the proportional
hazard model \cite{bib:miller1976least}.
This was overcome with the use of quantile regression
\cite{bib:buckley1979linear}, \cite{bib:peng2008survival}.
However, both the quantile regression AFT and the proportional hazards models
remain limited by the required linearity in the input covariates, and thus
neither can represent generic survival patterns.

Attempts have been made to incorporate neural networks into traditional
survival modelling frameworks.
In the Faraggi \& Simon proportional hazards model \cite{bib:faraggi1995neural}
the proportionality factor was modelled by the exponential of a single layer
perceptron, and more recently by a deep neural network in
\cite{bib:katzman2018deepsurv}.
A convolutional neural network was also used in \cite{bib:zhu2016deep} to model
survival functions based on image data.
Despite these recent advances, the proportional hazards framework still requires
the inclusion of time-dependencies to the proportionality factor.
A relevance vector machine extension of the AFT model was proposed by
\cite{bib:kiaee2015relevance}.
In this model, the survival time is restricted to a Weibull probability
distribution.
A neural network extension of the AFT is given by
\cite{bib:chapfuwa2018adversarial} in which the survival time is restricted to a
log-normal probability distribution.

Alternative machine learning methods that deconstruct the single estimation
problem into several sub-problems have also been explored.
First are cluster based methods that divide subjects into small groups
according to the values of the input attributes, and for each group uses a
method in which the survival function does not depend on the attributes
\cite{bib:ishwaran2008random,bib:ishwaran2014random,bib:chen2019nearest}.
Second are discrete time-interval models, in which each time-step corresponds to a
different classification problem \cite{bib:yu2011learning,bib:lee2018deephit,
bib:ren2019deep}.
The first approach relies on the availability of a large dataset, since only a
small part of the data is relevant to the estimation of each sub-model.
The second approach only computes the event probability for a finite number of
time points.

Survival data may also be modelled within a generative framework.
Here the probability distribution for the time-to-event is not directly
modelled, but is sampled instead. This sampling can be achieved by various methods, for example:
Gaussian processes \cite{bib:martino2011approximate,bib:joensuu2012risk,
	bib:fernandez2016gaussian,bib:alaa2017deep};
deep exponential families \cite{bib:ranganath2016deep,bib:miscouridou2018deep};
and generative adversarial nets \cite{bib:chapfuwa2018adversarial}.
However, these methods cannot be applied to problems that require explicit modeling of the survival
function.

In this paper, we propose a generic framework for integrating neural
networks within statistical survival modeling.
This is achieved with a parametric function of time whose parameters are
modelled as the output of a neural network.
This is referred to as a metaparametric framework that overcomes the problems identified previously, and for the first time:
\begin{enumerate}
\item provides a non-linear extension of the proportional hazards model;
\item includes a generic extension to time-dependencies of the proportionality
factor;
\item reduces the number of parameters required in discrete time-interval models by
aggregating data across infinite time intervals;
\item allows existing neural network survival analysis methods to fit into this
single framework.
\end{enumerate}
The proposed framework does not impose any ``a priori'' restriction to the type
of function that is being modelled.
It therefore extends to survival analysis one of the most important
capabilities that neural networks have in the regression and classification
domains, to represent any function without prior knowledge of its
structure.

The rest of this article is organized as follows.
Section \ref{sec:metaparametric_structure} gives the definition of the
metaparametric neural network structure and how it can be used to extend
existing neural network models for survival analysis.
Section \ref{sec:estimation} details the estimation of each type of
metaparametric neural network described in section
\ref{sec:metaparametric_structure}.
Sections \ref{sec:synthetic} and \ref{sec:experiments}  show the application of
the proposed framework to simulated and real-world survival datasets.
Finally, the conclusions of the study are given in Section \ref{sec:conclusion}.

\section{Metaparametric Structure in Neural Network Survival Models}
\label{sec:metaparametric_structure}
The goal of a survival model is to estimate the probability that an event will
happen in a given time interval.
The data used for this estimation is in the form $\mathcal{D} = \{
	\mathbf{x}_n; \allowbreak T_n; \allowbreak j_n; \allowbreak E_n
	| \allowbreak n \in \{1, \allowbreak 2, \allowbreak \dots,
	\allowbreak N\}\}$, where $\mathbf{x}_n$ are the input
attributes for subject $n$, $T_n$ is the time when subject $n$ experienced an
event or stopped being observed, $j_n$ is the type of event experienced by
subject $n$, and $E_n$ is $1$ if subject $n$ experienced an event at time $T_n$
or $0$ if the subject stopped being observed at time $T_n$ before experiencing any event.

The instantaneous probability of the event may be represented through the hazard
function: $\lambda(t,\mathbf{x})=f[T_{event}=t|\mathbf{x}, T_{event}\geq t]$.
Alternatively, the survival model can represent the probability of an event not
happening until time $t$, and is termed the survival function:
$S(t,\mathbf{x})=Pr[T_{event}\geq t|\mathbf{x}]$.
It is possible to alternate between representations using the cumulative hazard
function:
$\Lambda(t, \mathbf{x})=\int_{\nu=0}^{t}\lambda(t, \mathbf{x})\mathrm{d}\nu$;
where: $S(t,\mathbf{x})=\exp(-\Lambda(t,\mathbf{x}))$.
All representations are compatible with the possibility of the event never happening, in which case
$\lim_{t\rightarrow \infty}S(t,\mathbf{x})>0$.
However, both the survival function and the hazard function can only describe
single risk scenarios in which only one type of event is possible.
The hazard function can be extended to account for multiple competing risks in the form of the cause-specific hazard function:
$\lambda_j(t,\mathbf{x})=f[j,T_{event}=t|\mathbf{x}, T_{event}\geq t]$
\cite{bib:prentice1978analysis}.
In both single and competing risks scenarios, one or more time-dependent quantities are modelled as a function of the input covariates
$\mathbf{x}$.
A key challenge in building a generic neural network model for survival
analysis lies in the representation of these time-dependent quantities as outputs of
a neural network.

\subsection{Foundations of the metaparametric framework}
\label{sec:foundations}

In order to satisfy the requirements of a truly generic framework for survival analysis, the models must capture the nonlinear associations to the input variables by allowing parts that accept black box modelling, whilst satisfying the constraints relevant to the class of models. Here, a neural network is used to represent this black box nonlinear association within the hierarchical setting. This framework is termed the metaparametric neural network, and is defined as follows:

\begin{definition}
Let $\mathbf{\psi}(\mathbf{x},\mathbf{\theta})$ be a parametric neural network
with input variables $\mathbf{x}$ and parameters $\mathbf{\theta}$ and let
$g(\mathbf{y},\mathbf{\psi})$ be a parametric function of $\mathbf{y}$ with
parameters $\mathbf{\psi}$, where $\mathbf{y}$ is a set of input variables
disjunct from $\mathbf{x}$.
We define the metaparametric neural network (MNN)
$g(\mathbf{y},\mathbf{\psi}(\mathbf{x},\mathbf{\theta}))$ where the output of
$\mathbf{\psi}(\cdot)$ serves as the parameters of $g(\cdot)$.
This is a hierarchical model where the input variables are grouped into a set of
implicit variables $\mathbf{x}$ and another set of explicit variables
$\mathbf{y}$ that allows the outcome $g(\cdot)$ to be explicitly represented as a
function of $\mathbf{y}$ for any particular value of $\mathbf{x}$.
\end{definition}

In order to create a survival model with a MNN
structure, we choose time as the only explicit variable and $g(\cdot)$ to represent
the cause-specific hazard function. This structure is illustrated in Fig.
\ref{fig:metaparametric_neural_network}.

\tikzset{%
	opendot/.style={
    circle,
    fill=white,
    draw,
    outer sep=0pt,
    inner sep=1.5pt
  },
	block/.style  = {
		draw, thick, rectangle, minimum height = 3em,
		minimum width = 3em},
	sum/.style    = {draw, circle, node distance = 2cm}, 
	input/.style  = {coordinate}, 
	output/.style = {coordinate} 
}
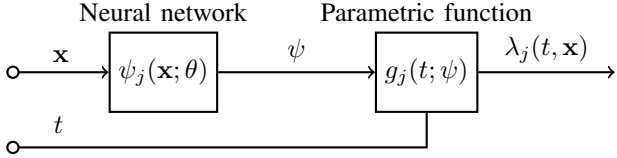
\begin{figure}[h]
\centering
\begin{tikzpicture}[auto, thick]

	\draw node at (0,0) [input, name=inputx] {};
	\draw node at (2,0.8) {Neural network};
	\draw node at (2,0) [block] (nn) {$\psi_{j}(\mathbf{x};\theta)$};

	\draw node at (0,-1) [input, name=inputt] {};
	\draw node at (5.5,0.8) {Parametric function};
	\draw node at (5.5,0) [block] (g) {$g_{j}(t;\psi)$};
	\draw[->] (nn) -- node{$\psi$} (g);
	\draw (inputt) -| node[xshift=-47mm,yshift=3mm]{$t$} (g);
	\draw[->] (inputx) -- node{$\mathbf{x}$} (nn);
    
	\draw node at (8,0) [output] (out) {};
	\draw[->] (g) -- node{$\lambda_j(t,\mathbf{x})$} (out);
    
	\draw node at (0,0) [opendot] {\Large};
	\draw node at (0,-1) [opendot] {\Large};

\end{tikzpicture}
\caption{Graphical description of a metaparametric neural network.}
\label{fig:metaparametric_neural_network}
\end{figure}

The MNN structure can be used as a generic framework
for survival analysis and current models can be described as specific cases of
it.
The existing neural network extensions of the proportional hazards model can be
cast in the metaparametric form.
This is achieved by making for each event type $j$:
$\lambda_j(t;\mathbf{x}) = \allowbreak g_j(t;\psi_j(\mathbf{x};\theta))
= \lambda_{0,j}(t)\exp(\psi_j(\mathbf{x};\theta))$,
where $\lambda_{0,j}(t)$ is the baseline hazard function for event type $j$ and $\psi_j(\mathbf{x};\theta)$ is the output of a neural network.

Similarly, the neural network versions of the AFT model can also be
expressed in a metaparametric form.
This is done by making $g(t;\psi(\mathbf{x};\theta))$ a log-normal probability
distribution with parameters
$\mu=\psi_1(\mathbf{x};\theta)$ and $\log\sigma=\psi_2(\mathbf{x};\theta)$, where
$\psi_{[1,2]}(\mathbf{x};\theta)$ are the outputs of a neural network.

The discrete time-interval models can also fit in a
metaparametric structure, with the use of a series of Kronecker delta functions.
This results in a cause-specific hazard function that is defined over a time
interval, as follows: $\lambda_j[\kappa;\mathbf{x}] = g_j[\kappa;\psi_{j,[0,\dots,K]}(\mathbf{x};\theta)] = \psi_{j,\kappa}(\mathbf{x};\theta)$, where $\kappa$ is the
index of a time interval and $\psi_{j,[0,\dots,K]}(\mathbf{x};\theta)$ are the
outputs of a neural network.

More importantly, the MNN framework can be used to formulate more generic
models.
This requires:
\begin{enumerate}
\item showing how the output of the MNN will describe the survival
	probability distribution, which is covered in section \ref{sec:generic};
\item making a choice of parametric function $g_{j}(t;\psi(\mathbf{x};\theta))$, which is covered
	in section \ref{sec:basis_functions}
\item estimating the parameters of the neural networks, which is covered in
	section \ref{sec:estimation}.
\end{enumerate}

\subsection{Generic metaparametric neural networks}
\label{sec:generic}
As shown in section \ref{sec:foundations}, the metaparametric structure provides
a formal generic framework for any neural network based survival model.
Here, we exploit this finding to derive novel extensions for all three classes of
survival models.

\subsubsection{Proportional hazards metaparametric neural network (PH-MNN)}
We define the PH-MNN with
the expression:
\begin{equation}
	\label{eq:ph_template}
	\lambda_j(t,\mathbf{x}) = \lambda_{0,j}(t)\omega_j(t,\mathbf{x})
\end{equation}
where $\lambda_{0,j}(t)$ is the cause-specific baseline hazard function and
$\omega_j(t,\mathbf{x})$ is the time-dependent hazard ratio, given by:
\begin{equation}
	\label{eq:ph_basis_extension}
	\omega_j(t,\mathbf{x}) = h\left(\sum_{k=1}^{K}
		\psi_{k,j}(\mathbf{x})\nu_{k}(t)\right)
\end{equation}
where $\nu_k(t)$ is a set of basis functions over time; $\psi_{k,j}(\mathbf{x})$
are outputs of a neural network; and $h(\cdot)$ is a strictly positive function.
Traditionally, $h(\cdot)$ is an exponential function.
The choice of the basis $\nu_k(t)$ and the function $h(\cdot)$ will strongly
influence the model estimation procedure and its computational requirements.
If the basis $\nu_k(t)$ is localized in time, being positive inside a finite
interval and null outside it, the following simplified structure is useful:
\begin{equation}
	\label{eq:ph_basis_inner_extension}
	\omega_j(t,\mathbf{x}) = \sum_{k=1}^{K}h\left(
		\psi_{k,j}(\mathbf{x})\right)\nu_{k}(t)
\end{equation}
Here, the time localization and non-negativity of the basis is required to
guarantee that $\omega_j(t,\mathbf{x})\geq 0$.
The variability in the amount of data for each type of event may dictate that we
choose a different basis set $\nu_k(t)$ for each event type $j$.

\subsubsection{Quantile regression metaparametric neural network (QR-MNN)}
We define the QR-MNN
quantile function as:
\begin{equation}
	\label{eq:aft}
	Q(\tau,\mathbf{x}) = \int_{u=0}^{-\log\tau} h\left(\sum_{k=1}^{K}
		\psi_{k}(\mathbf{x})\nu_{k}(u)\right) \mathrm{d}u
\end{equation}
where $Q(\tau,\mathbf{x}) = \inf\{t: 1-S(t|\mathbf{x})\geq \tau\}$.
The metaparametric formulation must respect the constraint that
$Q(\tau,\mathbf{x})$ should be strictly increasing with time.
A suitable basis set $\nu_k(t)$ should provide an analytical
expression for the integral in equation (\ref{eq:aft}).
Analogous to the PH-MNN model, the function $h(\cdot)$ can also be placed
inside the summation resulting in the following form:
\begin{equation}
	\label{eq:aft_inner}
	Q(\tau,\mathbf{x}) = \sum_{k=1}^{K}h\left(
		\psi_{k}(\mathbf{x})\right)
		\int_{u=0}^{-\log\tau} \nu_{k}(u)\mathrm{d}u
\end{equation}
This makes analytical integration more simple.
A competing risks extension can be achieved using a cause-specific quantile,
which we define as
$Q_j(\tau,\mathbf{x})=\inf\{t: 1-\exp(-\Lambda_j(t;\mathbf{x}))\geq \tau\}$, where
$\Lambda_j(t;\mathbf{x})=Pr[T_{event}<t;j|\mathbf{x}]$ is the cause-specific cumulative
hazard function.

Note that either the quantile function or its competing risks extension fully
specifies the event probability distribution and the correspondent hazard
function can be retrieved from it:
\begin{equation}
\label{eq:qr_conversion_competing}
	\lambda_j(t,\mathbf{x}) = -\frac{\mathrm{d}}{\mathrm{d}t}\log\left[1-Q_j^{-1}(t,\mathbf{x})\right]
\end{equation}

\subsubsection{Direct hazard metaparametric neural network (DH-MNN)}
We define the DH-MNN
as a continuous time extension of the discrete time-interval models.
This is achieved with the following formulation:
\begin{equation}
	\label{eq:direct}
	\lambda_j(t,\mathbf{x}) = h\left(\sum_{k=1}^{K}
		\psi_{k,j}(\mathbf{x})\nu_{k}(t)\right)
\end{equation}
where the function $h(\cdot)$ should be positive for the model to be
coherent, in the sense
that the hazard function is never negative.
This is a direct functional representation of the hazard function and; therefore, can be termed as a
direct hazard model.

An alternative formulation, as in the PH-MNN and QR-MNN, is:
\begin{equation}
	\label{eq:direct_inner}
	\lambda_j(t,\mathbf{x}) = \sum_{k=1}^{K}h\left(
		\psi_{k,j}(\mathbf{x})\right)\nu_{k}(t)
\end{equation}
where the basis set $\nu_k(t)$ should be positive and localized in time.

\subsubsection{General remarks}
In all the above models, an infinite set of basis functions can represent any square integrable function of time in a finite interval. Restricting the number of basis functions to be finite has an effect analogous to eliminating the high frequency components of the target function. In practice, a sufficient approximation accuracy to any function can be achieved by a suitable finite set of basis functions. This approach provides a continuous and smooth representation of the target function, whilst reducing the required number of basis functions, and consequently the risk of
overfitting. The extension of the proportional hazards model has the additional advantage of allowing the inclusion of high frequency components of the hazard function that are common to all values of $\mathbf{x}$, via the baseline hazard function.

\subsection{Choice of basis functions}
\label{sec:basis_functions}
This section contains a description of some of the possible choices of basis
functions for the generalizations proposed in
equations (\ref{eq:ph_basis_extension}) to (\ref{eq:direct_inner}).

\subsubsection{Piecewise constant basis functions}
Given a set of time knots $[\bar{T}_0, \bar{T}_1, ..., \bar{T}_K]$, the set
of basis functions
for a piecewise constant model will be:
\begin{equation}
	\nu_k(t) = \left\{\begin{split}
		1,\ & \bar{T}_{k-1}\leq t<\bar{T}_{k}\\
		0,\ & \text{otherwise}
	\end{split}\right.
\end{equation}
This set of basis functions is completely separated in time, simplifying
model computation.

For a PH-MNN model, this basis choice makes equations
(\ref{eq:ph_basis_extension}) and (\ref{eq:ph_basis_inner_extension}) equivalent
and removes the need to compute $h(\cdot)$ for each time point separately in the
objective function.
Also, this choice of basis functions allows analytical
conversion between different representations of the event probability
distribution for all MNN models, thereby reducing the computational cost of the estimation.

Although the computation is simpler than with other choices of basis functions,
a smooth transition between intervals cannot be achieved, with discontinuities
in the modelled hazard function despite the target function being smooth.

\subsubsection{Piecewise linear basis functions}
Given a set of time knots $[\bar{T}_0, \bar{T}_1, ..., \bar{T}_K]$, the set of basis functions
for a piecewise constant model will be:
\begin{equation}
	\nu_k(t) = \left\{\begin{split}
		(t-\bar{T}_{k-1})/(\bar{T}_{k}-\bar{T}_{k-1}),\ & \bar{T}_{k-1}\leq t<\bar{T}_{k}\\
		(\bar{T}_{k+1}-t)/(\bar{T}_{k+1}-\bar{T}_{k}),\ & \bar{T}_{k}\leq t<\bar{T}_{k+1}\\
		0,\ & \text{otherwise}
	\end{split}\right.
\end{equation}
In contrast to the piecewise constant models, the basis functions are continuous.
For the PH-MNN model, equations (\ref{eq:ph_basis_extension}) and
(\ref{eq:ph_basis_inner_extension}) are no longer equivalent.
Although both formulations are possible, (\ref{eq:ph_basis_inner_extension}) will have
smaller computational cost for estimation, as discussed in section
\ref{sec:estimation}.
The same is true of QR-MNN or DH-MNN models and computation will be simplified
with the use of equations
(\ref{eq:aft_inner}) and (\ref{eq:direct_inner}) respectively.
This choice of basis functions also
allow analytic conversion among different representations of the event
probability distribution, analogous to the piecewise constant basis functions.

\subsubsection{Other basis functions}
Other choices of basis functions are possible that make the resultant model
smoother than in the piecewise models.
These include the Fourier, polynomial, Chebyshev, and Legendre basis functions.
In this case, it is not possible to use the model formulations provided in
equations (\ref{eq:ph_basis_inner_extension}), (\ref{eq:aft_inner}) and
(\ref{eq:direct_inner}).
Instead, a similar effect is achieved by making $h(y) = y^2$ in equations
(\ref{eq:ph_basis_extension}), (\ref{eq:aft}) and (\ref{eq:direct}).
Given a set of unconstrained coefficients of $y$, a convolution property can be
used to compute a set of coefficients that will produce $y^2$ in the same basis
either in Fourier or in polynomial representations.
If the convolution property is used, $\Lambda_j(t,\mathbf{x})$ can be computed
analytically through the integration of each basis function individually.
For a QR-MNN model, the inverse of the quantile function cannot be computed
analytically, requiring a numerical approximation to be used.

\section{Estimation}
\label{sec:estimation}

\subsection{Proportional hazards metaparametric neural networks}
\label{sec:ph_generalization}

The original estimation method for the proportional hazards model is the
partial likelihood maximization \cite{bib:cox1972regression}, with competing
risks extensions proposed by
\cite{bib:prentice1978analysis,bib:larson1985mixture,bib:fine1999proportional}.
These estimators are compatible with time-dependent hazard ratios and
require no further development for implementation with the PH-MNN structure,
regardless of the different forms of the partial likelihood objective function.
However, special care is required in the implementation to avoid impractical
computational cost.
We show here the estimation procedure for the Cox partial likelihood estimator.
The same procedure can also be used for other objective functions.

The Cox partial log-likelihood is given by $\mathcal{L}=\sum_{n} \mathcal{L}_n$,
where:
\begin{equation}
	\mathcal{L}_n = \left[
		\log\omega(t,\mathbf{x}_n)
		-\log{\sum_{m=n}^{N}\omega(t,\mathbf{x}_m)}\right]E_n
\end{equation}
where $E_n$ indicates if an event has occurred to subject $n$ at time $T_n$.
For N subjects, the complexity of a training step is
$\mathcal{O}(N^2N_K+N(C_F+C_B))$, where $N_K$ is the number of basis functions,
and $C_F$ and $C_B$ are respectively the computational costs of feed-forward and
back-propagation in the chosen neural network architecture.
This is impractical for large datasets, and is avoided in traditional neural
networks by mini-batch approximation or by on-line training
\cite{bib:wilson2003general}.
Here, an extension of this technique is required since standard mini-batch
approximation would still lead to a computational cost that grows linearly with
$N$.
This is achieved by training the data with two independent sets of
mini batches:
\begin{enumerate}
\item \label{enum:batch} the first containing an arbitrary set of subjects with
	size $N_b$;
\item \label{enum:event} and the second containing only uncensored subjects with
	size $\tilde{N}_b$.
\end{enumerate}
The mini batch approximation of $\log{\sum_{m=n}^{N}\omega(t,\mathbf{x}_m)}$ is
achieved by replacing the summation with the average of $\omega(t,\mathbf{x}_m)$
for all $\mathbf{x}_m$ in the mini batch \ref{enum:batch}.
The approximation of $\mathcal{L}$ is given as the average of all
$\mathcal{L}_n$ in mini batch \ref{enum:event}.
For simplicity, we normalize the log-likelihood by the number of uncensored
subjects.
Note that for each subject in mini batch \ref{enum:event}, it is necessary to
make an independent estimation within mini batch \ref{enum:batch}.
Then, the cost of one training iteration becomes
$\mathcal{O}(N_K N_b \tilde{N}_b + (N_b+\tilde{N}_b)(C_F+C_B))$.

The estimation of the baseline hazard function requires consideration of time
variation.
The Kalbfleish \& Prentice estimator \cite{bib:kalbfleisch1980statistical} and
the Breslow estimator \cite{bib:breslow1974covariance} both provide an
analytical expression for the baseline hazard function, but assume a
time-invariant proportionality factor and a single risk.
However, Kalbfleish \& Prentice can be extended with the
cumulative cause-specific baseline hazard function taking the form:
\begin{equation}
	\label{eq:baseline_hazard_function}
	\Lambda_{0,j}(t) = \sum_{T_n<t;E_n=1;j_n=j}-\frac{
		\log\left[1-\frac{
			\omega_j(\mathbf{x}_n,T_n)}{
			\sum_{T_m \geq T_n}\omega_j(\mathbf{x}_m,T_n)}\right]}{
		\omega_j(\mathbf{x}_n,T_n)}
\end{equation}
Note that if the model is based in equation (\ref{eq:ph_basis_extension}),
the computational cost of estimating the survival probability for one single
subject after the model has been trained grows linearly with the training
dataset size.
If the model is based on equation (\ref{eq:ph_basis_inner_extension}), the
product in equation (\ref{eq:baseline_hazard_function}) can be rearranged so
that it only needs to be computed once and the computation of the survival
function for each new subject can be performed with complexity
$\mathcal{O}(C_F + \log N)$.

\subsection{Quantile regression metaparametric neural networks}
\label{sec:qr_estimation}

Estimation in the QR-MNN model is performed by maximizing its log-likelihood,
 given by:
\begin{equation}
	\label{eq:qr_likelihood}
	\mathcal{L} = \sum_{n=1}^{N}\left[\log(\lambda_{j_n}(\mathbf{x}_n,T_n))E_n-\sum_{j=1}^{J}\Lambda_{j}(\mathbf{x}_n,T_n)\right]
\end{equation}
where $\Lambda_{j}(\mathbf{x},t) = \int_{0}^{t}\lambda_{j}(\mathbf{x},\nu)\mathrm{d}\nu$
and the cause-specific hazard function $\lambda_{j}(\mathrm{x},t)$ can be
retrieved from the cause-specific quantile function in equation (\ref{eq:qr_conversion_competing}).
Here, standard mini-batch approximation can be performed.
Note that the estimation of this likelihood requires the computation of the
inverse of the quantile function, so estimation will be impacted by the choice
of basis functions as highlighted in section \ref{sec:basis_functions}.
If the basis function is chosen to be piecewise constant or piecewise linear, the inverse of the quantile
function can be computed analytically and the computational complexity of
training a single batch will be $\mathcal{O}(N_b (C_F+C_B))$, where $N_b$ is the size of
the mini-batch, and $C_F$ and $C_B$ are respectively the costs of feed-forward and
back-propagation in the chosen neural network architecture.

\subsection{Direct hazard metaparametric neural networks}
\label{sec:dt_generalization}
We estimate the DH-MNN  model by maximizing its log-likelihood, given by
equation (\ref{eq:qr_likelihood}) in section \ref{sec:qr_estimation}.
The computation of the likelihood is simplified if the version of the model in
equation (\ref{eq:direct_inner}) is used, since the integral can be
computed analytically.
Here, the standard mini-batch approximation can also be performed.
If the basis functions are chosen to be piecewise constant or piecewise linear, as detained in section \ref{sec:basis_functions}, the computational complexity of training a single batch will be $\mathcal{O}(N_b (C_F+C_B))$, as with the QR-MNN model.

\section{Application to synthetic data modeling}
\label{sec:synthetic}
In this section, we provide an example of the application of the proposed models
to estimate the cause-specific survival probability distribution in a synthetic
dataset.
The synthetic data used has two input covariate and two possible events, with
the cause-specific hazard function being:
$\lambda_1(t,\mathbf{x}) = 0.03 (1+0.5\cos(2\pi t/10))$ \allowbreak $\exp(
	\tan^{-1}(2x[0])\mathbbm{1}(t<5)+\tan^{-1}(2x[1])\mathbbm{1}(t>5))$;
$\lambda_2(t,\mathbf{x}) = 0.03 (1+0.5\sin(2\pi t/10)) \allowbreak \exp(
	\sin(x[1])\allowbreak\mathbbm{1}(t<5)+\sin(x[0])\mathbbm{1}(t>5))$,
where x[0] and x[1] are have independent normal distributions and
$\mathbbm{1}(\cdot)$ is the indicator function, which takes the value of $1$ when the argument is true and $0$ otherwise.

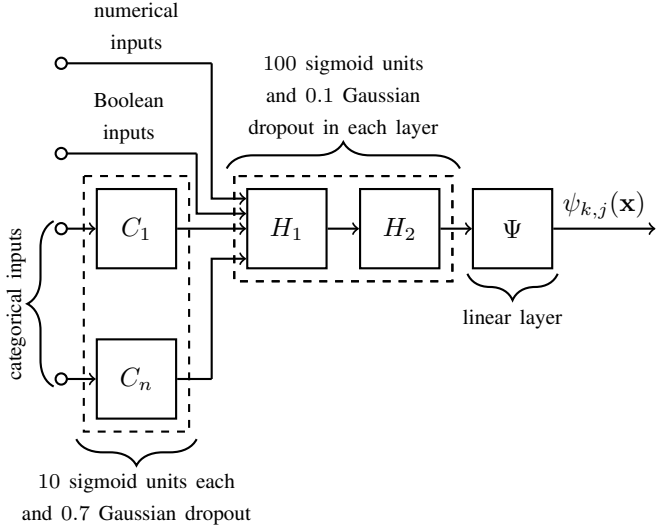
\begin{figure}[h]
\centering
\begin{tikzpicture}[auto, thick]

	\clip (-0.8,-5) rectangle (8, 2.1);
	\draw node at (0,1.2) [input, name=inputnum] {};
	\draw node at (0,0) [input, name=inputbool] {};
	\draw node at (0,-1) [input, name=inputcat1] {};
	\draw node at (0,-3) [input, name=inputcatn] {};
	\draw [decorate,decoration={brace,amplitude=10pt}](-0.1,-3.1) -- (-0.1,-0.9);
	\draw node at (-0.6,-2) [text width=4cm,align=center,rotate=90,inner sep=0,outer sep=0] (c)
		{\footnotesize categorical inputs};

	\draw node at (1,-1) [block] (c1) {$C_1$};
	\draw[->] (inputcat1) -- (c1.west);
	\draw node at (1,-3) [block] (cn) {$C_n$};
	\draw[->] (inputcatn) -- (cn.west);
	\draw[dashed] (0.3,-0.3) -- (0.3,-3.7) -- (1.7,-3.7)
		-- (1.7,-0.3) -- (0.3,-0.3);
	\draw [decorate,decoration={brace,amplitude=10pt}](1.8,-3.8) -- (0.2,-3.8);
	\draw node at (1,-4.6) [text width=4cm,align=center] (c) {\footnotesize $10$ sigmoid units each and $0.7$ Gaussian dropout};

	\draw node at (3,-1) [block] (h1) {$H_1$};
	\draw (inputnum) -- node[text width=1.5cm,align=center] {\footnotesize numerical inputs} (2,1.2);
	\draw[->] (2,1.2) |- ([yshift=4mm]h1.west);
	\draw (inputbool) -- node[text width=1.5cm,align=center] {\footnotesize Boolean inputs} (1.8,0);
	\draw[->] (1.8,0) |- ([yshift=2mm]h1.west);
	\draw[->] (c1) -- (h1.west);
	\draw (cn) -- (2,-3);
	\draw[->] (2,-3) |- ([yshift=-4mm]h1.west);
	\draw node at (4.5,-1) [block] (h2) {$H_2$};
	\draw[->] (h1) -- (h2);
	\draw[dashed] (2.3,-0.3) -- (2.3,-1.7) -- (5.2,-1.7)
		-- (5.2,-0.3) -- (2.3,-0.3);
	\draw [decorate,decoration={brace,amplitude=10pt}](2.2,-0.2) -- (5.3,-0.2);
	\draw node at (3.75,0.75) [text width=3cm,align=center] (c) {
		\footnotesize $100$ sigmoid units and $0.1$ Gaussian dropout in
		each layer};

	\draw node at (6,-1) [block] (psi) {$\Psi$};
	\draw[->] (h2) -- (psi);
	\draw[->] (psi) -- node {$\psi_{k,j}(\mathbf{x})$} (7.9,-1);
	\draw[dashed] (2.3,-0.3) -- (2.3,-1.7) -- (5.2,-1.7)
		-- (5.2,-0.3) -- (2.3,-0.3);
	\draw [decorate,decoration={brace,amplitude=10pt}](6.6,-1.6) -- (5.4,-1.6);
	\draw node at (6,-2.2) [text width=3cm,align=center] (c) {
		\footnotesize linear layer};
    
	\draw node at (0,1.2) [opendot] {\Large};
	\draw node at (0,0) [opendot] {\Large};
	\draw node at (0,-1) [opendot] {\Large};
	\draw node at (0,-3) [opendot] {\Large};

\end{tikzpicture}
\caption{Graphical description of the neural network structure applied in all
models.}
\label{fig:nn}
\end{figure}

The following models were compared:
\begin{itemize}
	\item \textbf{PH-MNN}:
		with piecewise linear basis functions and time knots equally distributed in
		intervals of 2.
	\item \textbf{QR-MNN}:
		with piecewise linear basis functions and quantile knots given by
		$\exp(-\Lambda_k)$ with $\Lambda_k \in \{0.01, 0.03, 0.06, 0.1, 0.2\}$.
	\item \textbf{DH-MNN}:
		with piecewise linear basis functions and time knots equally distributed in intervals
		of 2.
	\item \textbf{Cox}:
		the proportional hazard model \cite{bib:cox1972regression}
		with the baseline hazard function being estimated using the
		Kalbfleish-Prentice estimator
		\cite{bib:kalbfleisch1980statistical}.
		Competing risks were accounted for as in \cite{bib:prentice1978analysis}.
	\item \textbf{QR}: the quantile regression model with Lasso type penalty,
		as in \cite{bib:zheng2018high}.
	\item \textbf{DeepSurv}: a neural network adaptation
		of the Cox model \cite{bib:katzman2018deepsurv}, which is
		equivalent to a restricted version of the PH-MNN model with a
		single time constant basis function.
	\item \textbf{DeepHit}:
		a discrete time-interval model proposed in
		\cite{bib:lee2018deephit}, which can be viewed as a direct
		hazard model.
		Two different time discretization intervals of 2 and 0.1 were
		used to study the effect of a large discretization
		interval on the model.
		Being a discrete time model, the
		conversion between cumulative incidence function and
		cause-specific representations is only fully specified at the
		limit for an infinitely small discretization step.
		This might
		lead to a greater estimation error when a large
		discretization step is used.
\end{itemize}
In the PH-MNN, QR-MNN, DH-MNN, DeepSurv and DeepHit models, the same
structure was used for the neural network, which included Gaussian dropout
\cite{bib:srivastava2014dropout}.
This structure is described in Fig. \ref{fig:nn}.

\begin{figure}[h]
	\centering
	\includegraphics[scale=1]{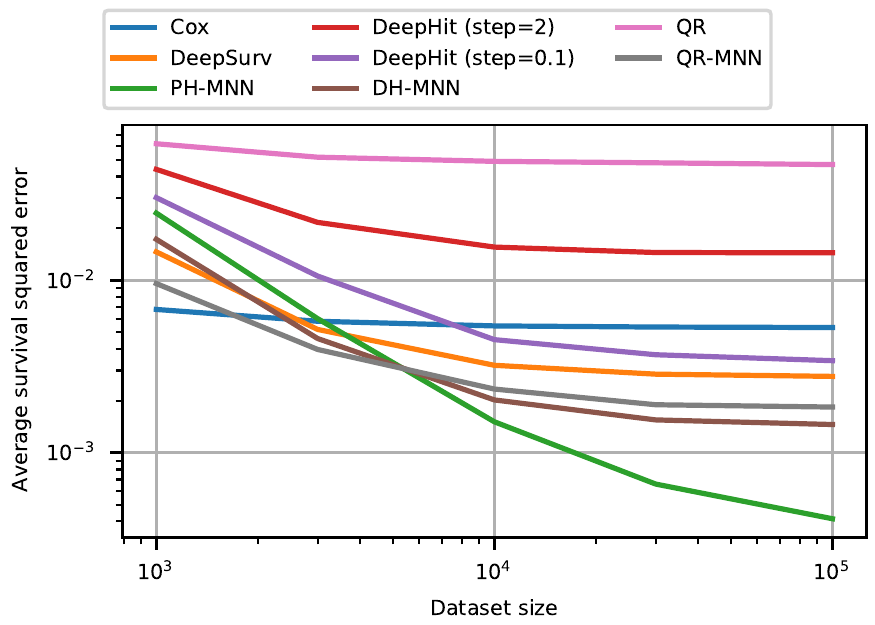}
	\caption{Averaged integrated squared error of the survival function
	for different dataset sizes.
	The results are the average of 100 independent models trained with
	independently generated datasets.}
	\label{fig:convergence}
\end{figure}

\begin{figure}[h]
	\centering
	\includegraphics[scale=1]{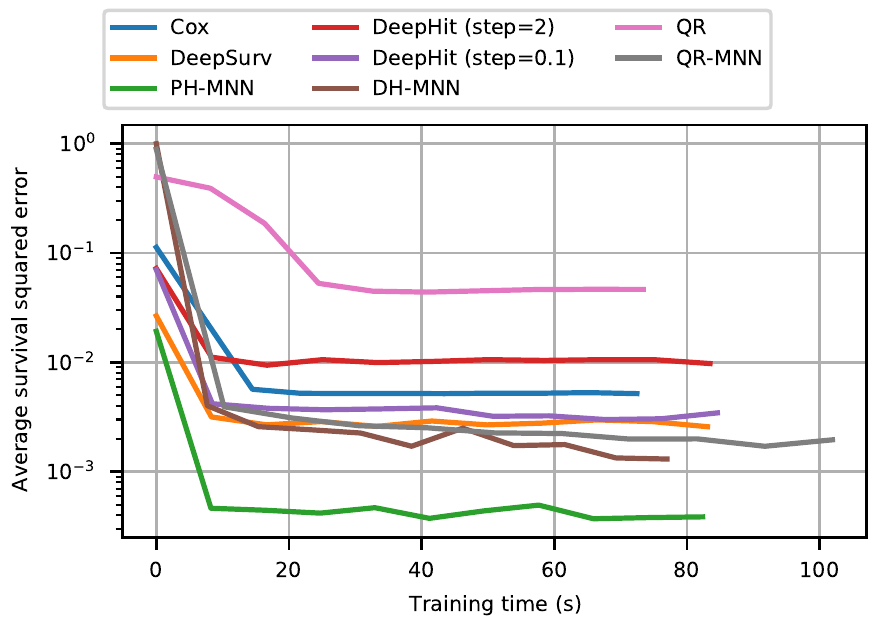}
	\caption{Averaged integrated squared error of the survival function
	over training time using a single synthetic dataset with 100000 data
	points.
	Computation was performed with a RTX 2070 graphics card and the models
	were implemented with TensorFlow 2.3.1.}
	\label{fig:training_time}
\end{figure}

\begin{figure}[h]
	\centering
	\includegraphics[scale=1]{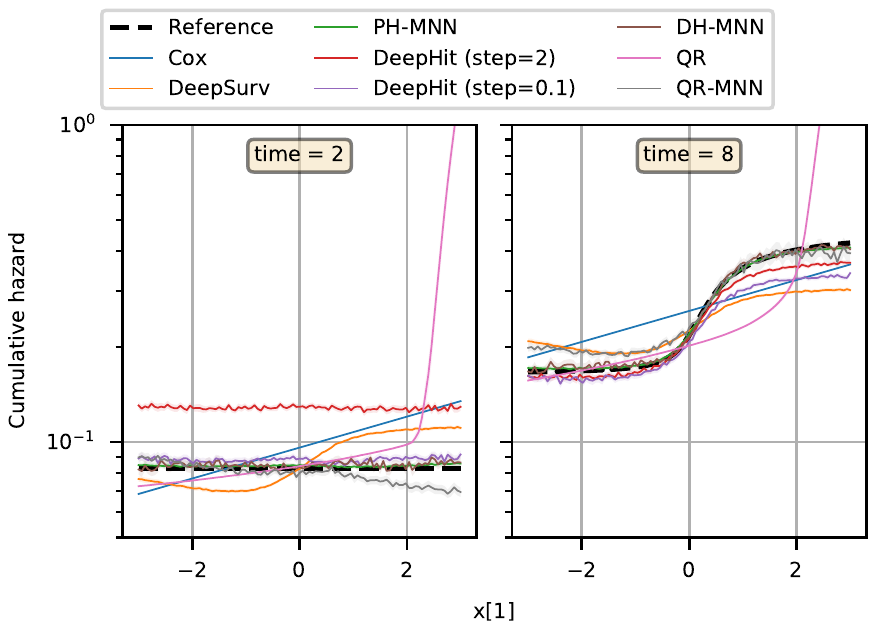}
	\caption{Cumulative cause-specific hazard function for event type $1$
	in the synthetic data as a function of time and variable $x[1]$ when
	$x[0]=0$.
	The values estimated are the averaged over 100 independent models
	trained with independently generated datasets, each one with 100000 data
	points.}
	\label{fig:synthetic_sensitivity}
\end{figure}

Fig. \ref{fig:convergence} shows how the averaged integrated squared error of
the survival function varies with training dataset size.
All of the MNN models performed better than previous state of the art.
With the exception of the PH-MNN, all models reached a saturation point where
the error ceases to improve at the same rate as a function of the dataset size.
This shows that the PH-MNN has more flexibility given the same number of
parameters as the other models, consistent with it's use of a nonparametric baseline hazard function.
Fig. \ref{fig:training_time} shows how the averaged integrated squared error of
the survival function evolves with model training time.
Although in neural networks the training time is flexible and comparing training
times of algorithms can be misleading, Fig. \ref{fig:training_time} shows that
all the proposed metaparametric neural network models have a shorter convergence
curve than their respective existing state-of-the-art models.
This means that the improvement achieved by the proposed models does not require
a higher computational time to be achieved.
Fig. \ref{fig:synthetic_sensitivity} shows how each model estimates the event
type $1$ in the synthetic data with $x[0]=0$ as a function of time and $x[1]$.
Note that all MNN models and also the DeepHit model are capable of representing
the nonlinearities and time-dependencies in the model with different accuracies,
as measured in Figs. \ref{fig:convergence} and \ref{fig:training_time}.
However, the DeepSurv, Cox and QR models are incapable of fully representing
the target probability distribution, and so they would never converge to the
underlying true probability distribution.

\section{Application to a clinical dataset}
\label{sec:experiments}
The proposed methodology was applied to the estimation of the risks of death and
revision surgery for patients who undergo hip replacement surgery, using data
collected by the National Joint Registry in the United Kingdom.
This dataset contains outcomes information from 1132875 hip replacement
surgeries performed from 2003 to 2019.
Here, modeling was restricted to procedures performed from April 2009 to March
2019.
Within this period, 855044 hip replacements were performed.
The data was filtered to include only surgeries with complete data and only
those where the reason for surgery was osteoarthritis, resulting in a total of
612914 procedures.

The observed population survival curves are shown in the Kaplan-Meier estimate
\cite{bib:kaplan1958nonparametric}.
The performance of the proposed MNN models, together with those of benchmark and
current state-of-the-art approaches were compared against the observed
Kaplan-Meier estimate.
The models used for comparison were the same as in section \ref{sec:synthetic}.
For the PH-MNN, the time knots used were 2, 4, and 7.
For the DH-MNN and DeepHit, time knots were equally distributed in intervals of
6 months.

\begin{figure}[h]
	\centering
	\includegraphics{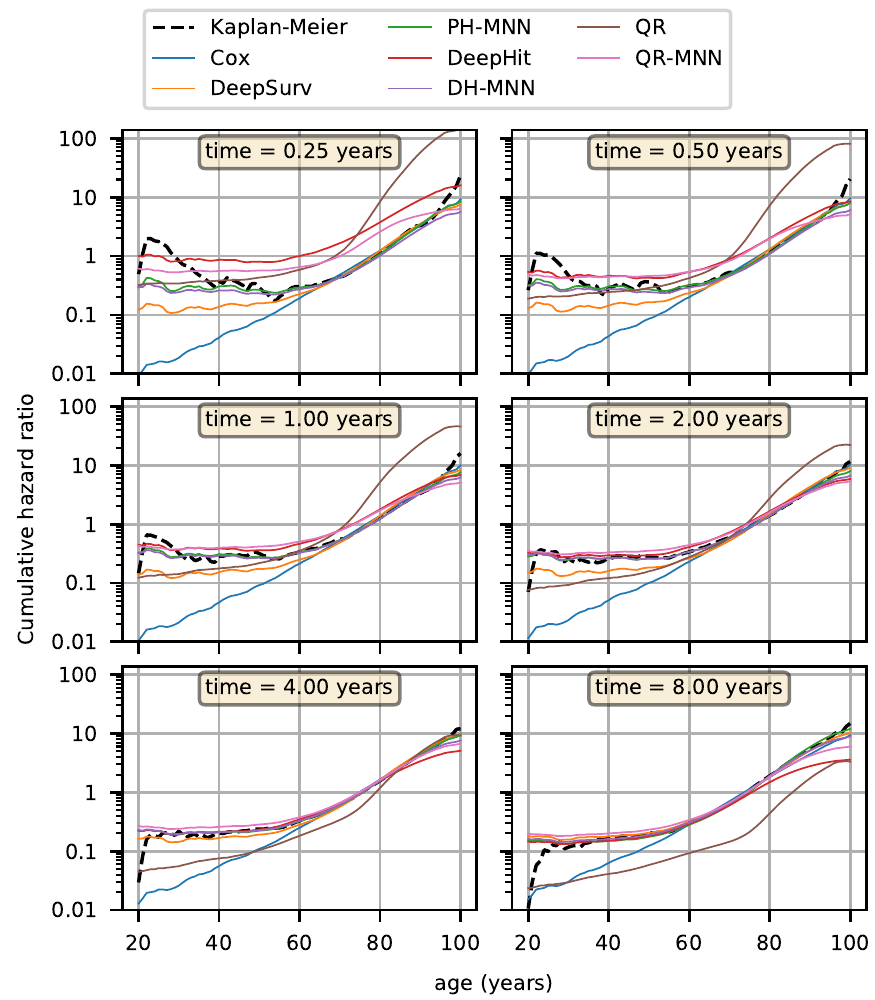}
	\caption{Estimated cumulative hazard ratio for the mortality risk
	marginalized as a function of the age.}
	\label{fig:mortality_by_age}
\end{figure}

\begin{figure}[h]
	\centering
	\includegraphics{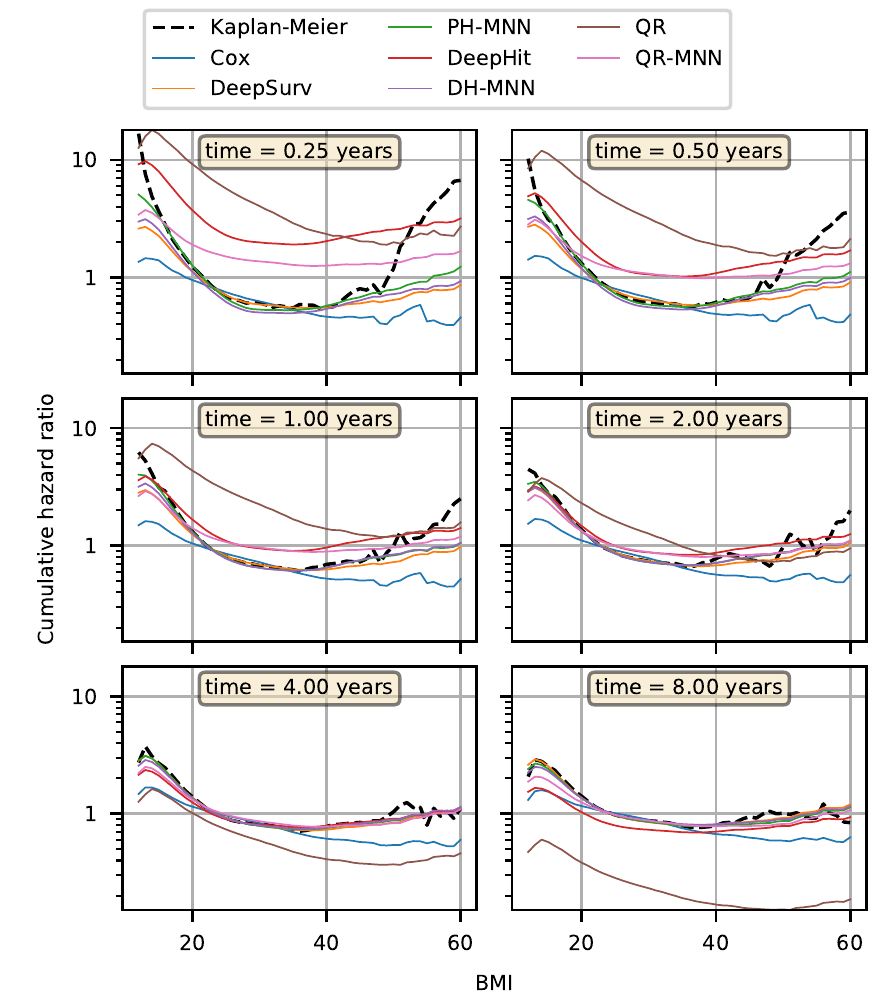}
	\caption{Estimated cumulative hazard ratio for the mortality risk
	marginalized as a function of the BMI.}
	\label{fig:mortality_by_bmi}
\end{figure}

\begin{figure}[h]
	\centering
	\includegraphics{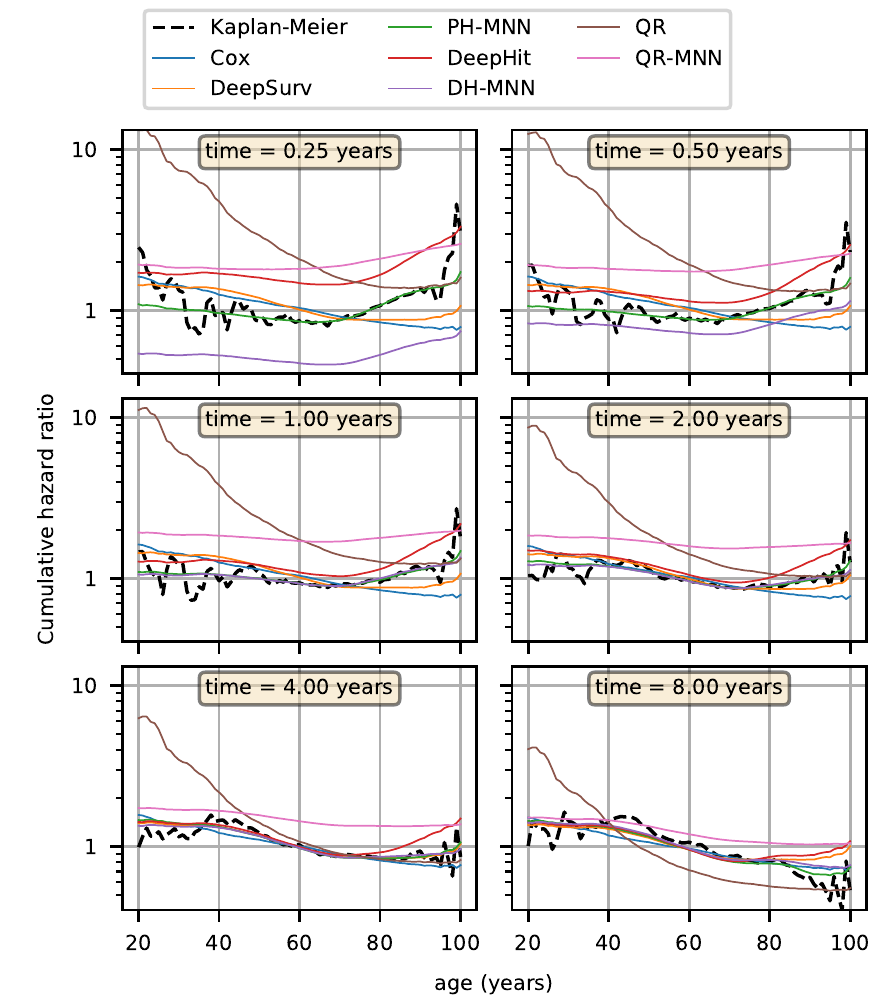}
	\caption{Estimated cumulative hazard ratio for the revision risk
	marginalized as a function of the age.}
	\label{fig:revision_by_age}
\end{figure}

\begin{figure}[h]
	\centering
	\includegraphics{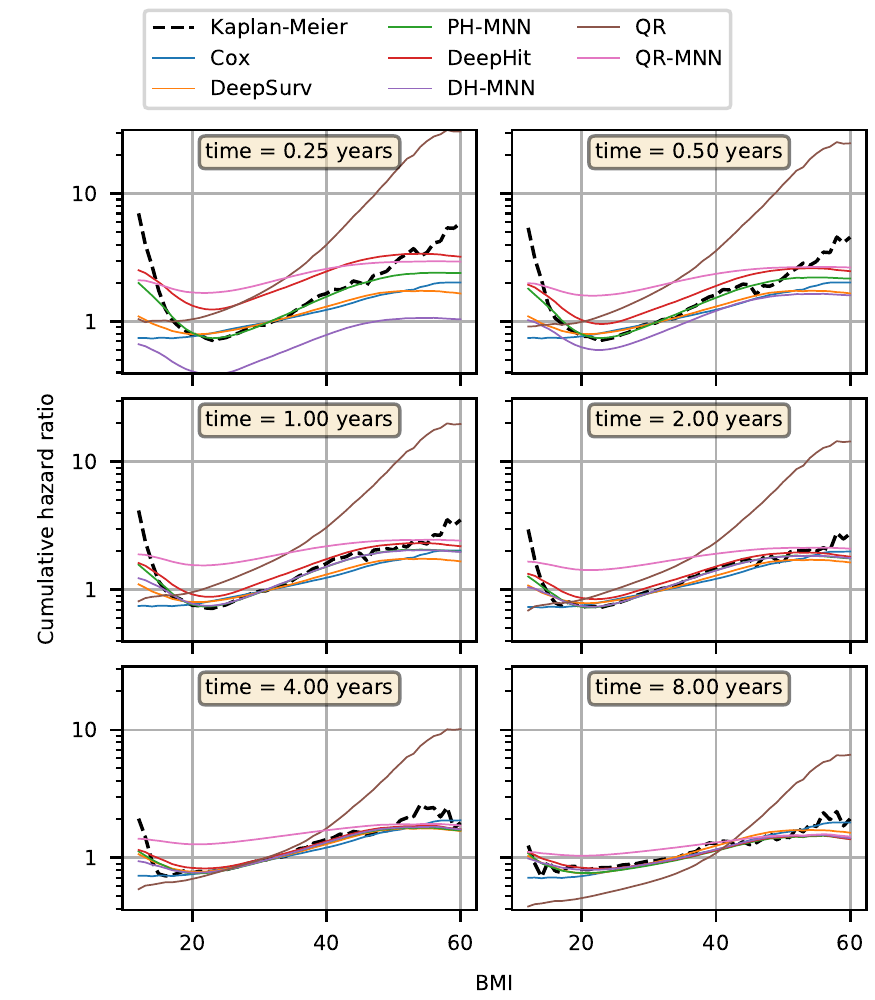}
	\caption{Estimated cumulative hazard ratio for the revision risk
	marginalized as a function of the BMI.}
	\label{fig:revision_by_bmi}
\end{figure}

The models were evaluated with plots of the estimated cumulative hazard ratio
(CHR) marginalized as a function of age and BMI.
We chose age and BMI as example predictor variables as they demonstrate a
nonlinear relationship with survival, which the proposed methods should be able
to capture.
The marginalized CHR estimation as a function of either the age or the BMI used
a sliding window with width equal to 4 in respective units and centered
successively in each target value, where:
\begin{itemize}
	\item the Kaplan-Meier estimate of the survival function within the
		window was performed, $S_{KM}(t|x\in \xi(w))$, where $\xi(w)$ is
		a window centered in $w$;
	\item the marginal model estimate within the window is computed as the
		average of the estimated survival function for each patient
		within the window, $S_{model}(t|x\in \xi(w))$;
	\item the Kaplan-Meier estimate was computed for the entire test
		population, $S_{KM}(t)$;
	\item the Kaplan-Meier estimation of the marginal CHR was given by:
		$\frac{\log(S_{KM}(t|x\in \xi(w)))}{\log(S_{KM}(t))}$.
	\item the model estimation of the marginal CHR was given by:
		$\frac{\log(S_{model}(t|x\in \xi(w)))}{\log(S_{KM}(t))}$.
\end{itemize}
This process was repeated 250 times, for each model in a group of 50 random
repetitions of 5-fold cross validation.
The results of the estimated marginal CHR as a function of age or BMI averaged
for all 250 runs are shown in Figs.
\ref{fig:mortality_by_age}-\ref{fig:revision_by_bmi}.
The results are evaluated according to the accuracy of representation
of nonlinearities, adaptability of the shape as a function of time, and
calibration.
These three aspects are captured by the root mean square error of the model
estimate of the log marginal CHR relative to the Kaplan-Meier estimate of the
same quantity.
For a given time $t$, this RMSE is given by:
\begin{align}
	&RMSE=
	\frac{\int p(x\in \xi(w))\log^2\left(\frac{\log(S_{model}(t|x\in \xi(w)))}{\log(S_{KM}(t|x\in \xi(w)))}\right)\mathrm{d}w}{\int p(x\in \xi(w))\mathrm{d}w}
\end{align}
where $p(x)$ is the population density inside the window $\xi(w)$ centered in $w$, and
$w$ is either the age or the BMI.
This RMSE represents an integrated measure of two factors:
first, the difference between the relationship of the model estimate and the
observed data as a function of the attribute;
and second the systematic bias between the two that is common for all values of
the attribute.
By estimating a bias that will minimize this RMSE, the two components can be
identified as the unbiased RMSE (URMSE) and the bias.
The model were evaluated through the computation of the RMSE, URMSE and absolute
bias in the time interval from 6 months to 8 years with steps of 1 month.
Tables \ref{tab:ph_table}, \ref{tab:dh_table} and \ref{tab:qr_table} present for
each type of model the maximum value over time of each evaluation criteria with
their $95\%$ confidence interval.
To highlight the improvement achieved with the metaparametric neural network
structure, the results were grouped by type of model.
For the direct hazards models, evaluation was performed in 6 months intervals to
allow a fair comparison between both the discrete and continuous-time models.

\begin{table}[h]
\renewcommand{\arraystretch}{1.3}
\setlength{\tabcolsep}{3pt}
\caption{Maximum value over time of each error component in proportional hazards models}
\label{tab:ph_table}
\centering
\begin{tabular}{|c|c|c|c|c|}
\hline
 & & Cox & DeepSurv & PH-MNN \\ \hline
\hline \multirow{3}{1.1cm}{\centering Revision by Age }
 & RMSE & $0.225 \pm 0.002$ & $0.213 \pm 0.002$ & $\mathbf{0.154 \pm 0.003}$ \\ \cline{2-5}
 & URMSE & $0.221 \pm 0.003$ & $0.209 \pm 0.002$ & $\mathbf{0.146 \pm 0.002}$ \\ \cline{2-5}
 & abs. bias & $0.060 \pm 0.003$ & $\mathbf{0.057 \pm 0.003}$ & $0.070 \pm 0.002$ \\ \hline
\hline \multirow{3}{1.1cm}{\centering Mortality by Age }
 & RMSE & $0.567 \pm 0.002$ & $0.313 \pm 0.004$ & $\mathbf{0.189 \pm 0.004}$ \\ \cline{2-5}
 & URMSE & $0.522 \pm 0.002$ & $0.279 \pm 0.003$ & $\mathbf{0.181 \pm 0.004}$ \\ \cline{2-5}
 & abs. bias & $0.243 \pm 0.002$ & $0.159 \pm 0.003$ & $\mathbf{0.071 \pm 0.004}$ \\ \hline
\hline \multirow{3}{1.1cm}{\centering Revision by BMI }
 & RMSE & $0.148 \pm 0.002$ & $0.140 \pm 0.002$ & $\mathbf{0.124 \pm 0.002}$ \\ \cline{2-5}
 & URMSE & $0.140 \pm 0.001$ & $0.130 \pm 0.002$ & $\mathbf{0.107 \pm 0.002}$ \\ \cline{2-5}
 & abs. bias & $0.074 \pm 0.003$ & $\mathbf{0.068 \pm 0.003}$ & $0.081 \pm 0.002$ \\ \hline
\hline \multirow{3}{1.1cm}{\centering Mortality by BMI }
 & RMSE & $0.197 \pm 0.002$ & $0.135 \pm 0.002$ & $\mathbf{0.133 \pm 0.003}$ \\ \cline{2-5}
 & URMSE & $0.193 \pm 0.001$ & $0.127 \pm 0.002$ & $\mathbf{0.121 \pm 0.003}$ \\ \cline{2-5}
 & abs. bias & $0.053 \pm 0.003$ & $\mathbf{0.052 \pm 0.003}$ & $0.061 \pm 0.003$ \\ \hline
\end{tabular}
\end{table}

\begin{table}[h]
\renewcommand{\arraystretch}{1.3}
\caption{Maximum value over time of each error component in direct hazards models}
\label{tab:dh_table}
\centering
\begin{tabular}{|c|c|c|c|}
\hline
 & & DeepHit & DH-MNN \\ \hline
\hline \multirow{3}{1.1cm}{\centering Revision by Age }
 & RMSE & $0.456 \pm 0.003$ & $\mathbf{0.260 \pm 0.004}$ \\ \cline{2-4}
 & URMSE & $0.162 \pm 0.002$ & $\mathbf{0.148 \pm 0.002}$ \\ \cline{2-4}
 & abs. bias & $0.430 \pm 0.002$ & $\mathbf{0.217 \pm 0.004}$ \\ \hline
\hline \multirow{3}{1.1cm}{\centering Mortality by Age }
 & RMSE & $1.008 \pm 0.004$ & $\mathbf{0.228 \pm 0.005}$ \\ \cline{2-4}
 & URMSE & $0.194 \pm 0.003$ & $\mathbf{0.180 \pm 0.004}$ \\ \cline{2-4}
 & abs. bias & $0.990 \pm 0.004$ & $\mathbf{0.144 \pm 0.005}$ \\ \hline
\hline \multirow{3}{1.1cm}{\centering Revision by BMI }
 & RMSE & $0.444 \pm 0.002$ & $\mathbf{0.242 \pm 0.003}$ \\ \cline{2-4}
 & URMSE & $0.111 \pm 0.002$ & $\mathbf{0.107 \pm 0.002}$ \\ \cline{2-4}
 & abs. bias & $0.431 \pm 0.002$ & $\mathbf{0.218 \pm 0.004}$ \\ \hline
\hline \multirow{3}{1.1cm}{\centering Mortality by BMI }
 & RMSE & $0.990 \pm 0.003$ & $\mathbf{0.184 \pm 0.005}$ \\ \cline{2-4}
 & URMSE & $0.133 \pm 0.002$ & $\mathbf{0.120 \pm 0.002}$ \\ \cline{2-4}
 & abs. bias & $0.981 \pm 0.003$ & $\mathbf{0.140 \pm 0.005}$ \\ \hline
\end{tabular}
\end{table}

\begin{table}[h]
\renewcommand{\arraystretch}{1.3}
\caption{Maximum value over time of each error component in quantile regression models}
\label{tab:qr_table}
\centering
\begin{tabular}{|c|c|c|c|}
\hline
 & & QR & QR-MNN \\ \hline
\hline \multirow{3}{1.1cm}{\centering Revision by Age }
 & RMSE & $0.697 \pm 0.005$ & $\mathbf{0.596 \pm 0.052}$ \\ \cline{2-4}
 & URMSE & $0.341 \pm 0.003$ & $\mathbf{0.171 \pm 0.005}$ \\ \cline{2-4}
 & abs. bias & $0.609 \pm 0.005$ & $\mathbf{0.570 \pm 0.053}$ \\ \hline
\hline \multirow{3}{1.1cm}{\centering Mortality by Age }
 & RMSE & $1.379 \pm 0.005$ & $\mathbf{0.576 \pm 0.044}$ \\ \cline{2-4}
 & URMSE & $0.730 \pm 0.006$ & $\mathbf{0.280 \pm 0.011}$ \\ \cline{2-4}
 & abs. bias & $1.364 \pm 0.005$ & $\mathbf{0.500 \pm 0.045}$ \\ \hline
\hline \multirow{3}{1.1cm}{\centering Revision by BMI }
 & RMSE & $\mathbf{0.595 \pm 0.006}$ & $0.598 \pm 0.054$ \\ \cline{2-4}
 & URMSE & $0.187 \pm 0.002$ & $\mathbf{0.154 \pm 0.007}$ \\ \cline{2-4}
 & abs. bias & $\mathbf{0.566 \pm 0.006}$ & $0.578 \pm 0.054$ \\ \hline
\hline \multirow{3}{1.1cm}{\centering Mortality by BMI }
 & RMSE & $1.682 \pm 0.007$ & $\mathbf{0.557 \pm 0.039}$ \\ \cline{2-4}
 & URMSE & $0.237 \pm 0.003$ & $\mathbf{0.176 \pm 0.005}$ \\ \cline{2-4}
 & abs. bias & $1.665 \pm 0.007$ & $\mathbf{0.515 \pm 0.042}$ \\ \hline
\end{tabular}
\end{table}

The PH-MNN, DH-MNN, QR-MNN, DeepSurv and DeepHit models captured the nonlinearities,
while the Cox did not.
The QR model partially captured some nonlinearities through the variation of
coefficients with the quantile, but they were not entirely captured since this
variation is shared to represent both nonlinearities and time variations.
This can be seen in the figures and is reflected by a smaller URMSE for the
neural network models in most cases. The nonlinearities of the CHR could be adapted as a function of time for all the metaparametric neural networks and for the DeepHit model.
The model structure for the others does not permit this variation of nonlinearities over time.

In the case of the proportional hazards models, the PH-MNN overall performance
measured by the RMSE was better than the established methods.
When this RMSE measure is broken down into its components, URMSE and absolute bias, the DeepSurv model had a slightly smaller bias. However the PH-MNN bias was still small and stable across the different risks and attributes. For the direct hazards models, the DH-MNN overall performance, URMS and bias were all consistently better than DeepHit. Finally, for the quantile regression models, the QR-MNN model overall
performance was also consistently better than the QR method, apart from revision by BMI in which the two methods were equivalent.

\section{Conclusion}
\label{sec:conclusion}

In this article, we propose a novel and generic framework for incorporating
machine learning into survival modeling that resolves the established
limitations of neural network application to this field.
This MNN framework results in a structure that encompasses existing neural
network models.
This framework enables the generic representation of any survival probability
distribution without prior knowledge of its functional form.
The MNN framework can be applied to both parametric and semi-parametric modeling
scenarios providing unification in this domain.
Special instances of this class of models were formulated based on three
different heritage structures: the proportional hazards, the quantile regression
and the direct hazard.
Both conventional and the novel framework models were evaluated using both a
synthetic and a large real-world dataset, with best overall fit to the observed
data being demonstrated by the proposed MNN framework.

\section*{Acknowledgment}

We thank the patients and staff of all the hospitals who have contributed
data to the National Joint Registry.
We are grateful to the Healthcare Quality Improvement Partnership (HQIP),
the National Joint Registry Steering Committee (NJRSC), and staff at the
NJR Centre for facilitating this work.
The views expressed represent those of the authors and do not necessarily
reflect those of the NJRSC or HQIP who do not vouch for how the information
is presented.




\bibliographystyle{IEEEtran}
\bibliography{IEEEabrv,bibliography}
\end{document}